\documentclass[10pt,journal,compsoc]{IEEEtran}
\ifCLASSOPTIONcompsoc
  \usepackage[nocompress]{cite}
\else
  \usepackage{cite}
\fi
\ifCLASSINFOpdf
\else
\fi
\newcommand\MYhyperrefoptions{bookmarks=true,bookmarksnumbered=true,
pdfpagemode={UseOutlines},plainpages=false,pdfpagelabels=true,
colorlinks=true,linkcolor={black},citecolor={black},urlcolor={black},
pdftitle={TransMatting: Tri-token Equipped Transformer Model for Image Matting},%<!CHANGE!
pdfsubject={TransMatting},%<!CHANGE!
pdfauthor={Huanqia Cai, Fanglei Xue, Lele Xu, Lili Guo},%<!CHANGE!
pdfkeywords={Tri-token, Vision Transformer, Image Matting, Deep Learning}}%<^!CHANGE!
\usepackage{booktabs}
\usepackage{multirow}
\ifCLASSINFOpdf
\usepackage[\MYhyperrefoptions,pdftex]{hyperref}
\else
\usepackage[\MYhyperrefoptions,breaklinks=true,dvips]{hyperref}
\usepackage{breakurl}
\fi
\newcommand{\Ourds}{Transparent-460}
\usepackage{graphicx}
\usepackage{listings}
\usepackage{mathrsfs}
\usepackage{amsmath,amssymb}
\usepackage{colortbl}
\usepackage{xcolor}
\definecolor{codegreen}{rgb}{0,0.6,0}
\definecolor{codegray}{rgb}{0.5,0.5,0.5}
\definecolor{codepurple}{rgb}{0.58,0,0.82}
\definecolor{backcolour}{rgb}{255,255,255}

\lstdefinestyle{mystyle}{
  backgroundcolor=\color{backcolour},   commentstyle=\color{codegreen},
  keywordstyle=\color{magenta},
  numberstyle=\tiny\color{codegray},
  stringstyle=\color{codepurple},
  basicstyle=\ttfamily\scriptsize,
  breakatwhitespace=false,         
  breaklines=true,                 
  captionpos=b,                    
  keepspaces=true,                 
  numbers=left,                    
  numbersep=5pt,                  
  showspaces=false,                
  showstringspaces=false,
  showtabs=false,                  
  tabsize=2
}
\usepackage{listings}
\usepackage{algorithm} 

\begin{document}
\title{TransMatting: Tri-token Equipped Transformer Model for Image Matting}


\author{Huanqia~Cai,
        Fanglei Xue,
        Lele Xu, 
        and Lili Guo
        %John~Doe,~\IEEEmembership{Fellow,~OSA,}
        %and~Jane~Doe,~\IEEEmembership{Life~Fellow,~IEEE}% <-this % stops a space
\IEEEcompsocitemizethanks{
\IEEEcompsocthanksitem Huanqia Cai and Fanglei Xue are with the University of Chinese Academy of Sciences, Beijing, China. E-mail: \{caihuanqia19, xuefanglei19\}@mails.ucas.ac.cn.
\IEEEcompsocthanksitem Huanqia Cai, Fanglei Xue, Lele Xu and Lili Guo are with the Key Laboratory of Space Utilization, Technology and Engineering Center for Space Utilization, Chinese Academy of Sciences, Beijing, China. Email:  \{xulele, guolili\}@csu.ac.cn. 
\IEEEcompsocthanksitem Huanqia Cai and Fanglei Xue contribute equally to this paper. Corresponding author: Lele Xu.
\IEEEcompsocthanksitem This work was supported by the National Natural Science Foundation of China under Grant 61901454. Project page: \url{https://github.com/AceCHQ/TransMatting}.
}

% \IEEEcompsocthanksitem M. Shell was with the Department
% of Electrical and Computer Engineering, Georgia Institute of Technology, Atlanta,
% GA, 30332.\protect\\
% E-mail: see http://www.michaelshell.org/contact.html
%\IEEEcompsocthanksitem J. Doe and J. Doe are with Anonymous University.}% <-this % stops a space
% \thanks{Manuscript received April 19, 2005; revised August 26, 2015.}
}


\markboth{Journal of \LaTeX\ Class Files,~Vol.~14, No.~8, August~2015}%
{Shell \MakeLowercase{\textit{et al.}}: Bare Advanced Demo of IEEEtran.cls for IEEE Computer Society Journals}



\IEEEtitleabstractindextext{%
\begin{abstract}
Image matting aims to predict alpha values of elaborate uncertainty areas of natural images, like hairs, smoke, and spider web. However, existing methods perform poorly when faced with highly transparent foreground objects due to the large area of uncertainty to predict and the small receptive field of convolutional networks. To address this issue, we propose a Transformer-based network (TransMatting) to model transparent objects with long-range features and collect a high-resolution matting dataset of transparent objects (Transparent-460) for performance evaluation. Specifically, to utilize semantic information in the trimap flexibly and effectively, we also redesign the trimap as three learnable tokens, named tri-token. Both Transformer and convolution matting models could benefit from our proposed tri-token design. By replacing the traditional trimap concatenation strategy with our tri-token, existing matting methods could achieve about 10\% improvement in SAD and 20\% in MSE. Equipped with the new tri-token design, our proposed TransMatting outperforms current state-of-the-art methods on several popular matting benchmarks and our newly collected Transparent-460.

\end{abstract}

}

\maketitle

% is not selected <OR> if conference mode is selected - because compsoc
% conference papers position the abstract like regular (non-compsoc)
\IEEEdisplaynontitleabstractindextext

% \fi
\IEEEpeerreviewmaketitle

\ifCLASSOPTIONcompsoc
\IEEEraisesectionheading{\section{Introduction}\label{sec:introduction}}
\else
\section{Introduction}
\label{sec:introduction}
\fi

\IEEEPARstart{I}{mage} matting is a technique to separate the foreground object from the background in an image by predicting a precise alpha matte. It has been widely used in many applications, such as image and video editing, background replacement, and virtual reality \cite{chen2013knn,xu2017deep,levin2007closed}. Image matting assumes that every pixel in the image $I$ is a linear combination of the foreground object $F$ and the background $B$ by an alpha matte $\alpha$:

\begin{equation}
    I = \alpha F + (1-\alpha)B,   \alpha \in [0, 1]
\end{equation}

As only the image $I$ is known in this equation, image matting is an ill-posed problem. So many existing methods \cite{chen2013knn,levin2007closed,sun2004poisson,wang2007optimized,xu2017deep,lu2019indices,li2020natural} take a trimap as an auxiliary input. The trimap is a single-channel image with three unique values: 255, 0, and 128. These values segment the original image into three parts: known foreground, known background and unknown areas.

Most traditional methods, including sampling-based \cite{berman2000method,chuang2001bayesian,wang2007optimized,gastal2010shared,he2011global,shahrian2013improving} and propagation-based methods \cite{chen2013knn,lee2011nonlocal,sun2004poisson,levin2007closed}, utilize the known area samples to find candidate colors or propagate the known alpha value. They heavily rely on information from known areas, especially the known foreground areas. Recently, learning-based methods
greatly improve image matting performance, but they still need specific information from known areas to predict unknown areas~\cite{liu2021long, li2020natural}.  
However, according to \cite{liu2021long}, more than 50\% pixels in the unknown areas cannot be correlated to pixels in the known regions due to the limited reception field of deep learning methods.
Furthermore, some highly transparent objects (\textit{e.g.}, glass, bonfires, plastic bags, etc.) and even non-salient objects (\textit{e.g.}, web, smoke, water drops, etc.) have meticulous interiors with most pixels belonging to the unknown regions \cite{li2021deep}.
It is very challenging for existing models to learn long-range features with little known information.
Tab.~\ref{tab:TTTP} illustrates the performance of some state-of-the-art (SOTA) methods on transparent totally (TT) and transparent partially (TP) objects separately on the Composition-1k test set. We could observe that the results of TT are much worse than TP, indicating that TT objects are the key to improving the overall evaluation performance.

To address this issue, we make the first attempt to introduce Vision Transformer (ViT) \cite{dosovitskiy2020image} to extract features with a large receptive field. The Transformer model is first proposed in natural language processing (NLP) and has achieved great performance in computer vision tasks, such as classification\cite{dosovitskiy2020image,touvron2021training,liu2021swin}, segmentation \cite{zheng2021RethinkingSemantic,lu2021SimplerBetter}, and detection \cite{carion2020DETR,yang2021focal}. It mainly consists of multi-head self-attention (MHSA) and multi-layer perception modules. Different from the convolution, the MHSA module could mine information in a global scope. Thus, the ViT model could learn global semantic features of the foreground object with high-level position relevance. To further help the model integrate the low-level appearance features (\textit{e.g.}, texture) with high-level semantic features (\textit{e.g.}, shape), a Multi-scale Global-guided Fusion (MGF) module is proposed. The MGF takes three adjacent scales of features as input, uses the non-background mask to guide the low-level feature, and employs the high-level feature to guide the information integration. With this new MGF module, only foreground features could be transmitted to the decoder, reducing the influence of background noises.

On the other hand, most current trimap-based methods follow DIM \cite{xu2017deep} to treat the trimap as an image and concatenate it with the RGB image before feeding into the deep model. This brings the same weakness: insufficient long-range relationships extracted from the trimap. Meanwhile, compared with the RGB image, the information in trimap is very sparse and has some high-level positional relevance \cite{liu2021tripartite}. Most areas in the trimap have the same value, making convolution neural networks with small kernels inefficient in extracting features from trimaps.
Different from the traditional manner, we propose a new form of trimap named \textbf{tri-token} with inspirations from the \texttt{[cls]} token in ViT. We utilize three tokens to represent three kinds of semantic information in the trimap: foreground, background, and unknown area. Based on the proposed tri-token, we could directly introduce these semantic information into the self-attention mechanism. 
The Transformer module could identify which features are from the known areas and which are from the unknown areas without any range limitation. In addition, profiting from the semantic-representation capability of our proposed tri-token, we could introduce trimap information to any depth of both Transformer and convolutional neural networks (CNNs). Experiments on various SOTA methods verify that the performance could be further improved by only replacing the traditional trimap with our tri-token.

\begin{table}
\setlength\tabcolsep{4pt}
\centering
\caption{Performance of transparent totally (TT) and transparent partially (TP) objects on the Composition-1k test set. TT objects are those whose entire image is highly transparent, while TP denotes objects with the significant known foreground.}
\label{tab:TTTP}
\begin{tabular}{ccccccc}
\toprule
\multirow{2}{*}{Methods} & \multicolumn{3}{c}{MSE$\downarrow$} & \multicolumn{3}{c}{SAD$\downarrow$} \\ \cline{2-7} 
                         & TT     & TP    & TT+TP  & TT     & TP     & TT+TP \\ \hline
IndexNet~\cite{lu2019indices}  & 22.87  & 8.9   & 13     & 110.3  & 18.08  & 45.8  \\
GCAMatting~\cite{li2020natural} & 15.89  & 6.2   & 9.1    & 85.72  & 13.68  & 35.3  \\
MGMatting~\cite{yu2021mask}  & 13.01  & 4.65  & 7.18   & 77.88  & 11.87  & 31.76 \\ \midrule
TransMattingV1                     & 7.49   & 3.40   & 4.58   & 59.37  & 10.35  & 24.96 \\ 
TransMattingV2                     & \textbf{6.94}   & \textbf{3.26}   & \textbf{4.37}   & \textbf{55.47}  & \textbf{10.25}  & \textbf{23.82} \\ \bottomrule
\end{tabular}
\end{table}

Besides, there has not been any test bed for images with transparent or non-salient foreground objects. Previous datasets mainly focus on salient and opaque foregrounds, like animals \cite{li2020endtoend} and portraits \cite{shen2016deep,liu2021tripartite}, which have significantly been investigated. To further help the community to dig into the transparent and non-salient cases, we collect 460 high-solution natural images with large unknown areas and manually label their alpha mattes.

To summarize, our main contributions are as follows:
\begin{enumerate}
\item We propose a Transformer-based model (TransMatting) with a big receptive field to help extract global features from images, especially for transparent objects. An MGF module is also introduced to guide the integration of multi-scale features with these global features.

\item We redesign the trimap in a semantic tri-token manner to directly integrate with deep features. The tri-token could introduce trimap information to image features flexibility by reducing the domain gap between them.

\item We build a high-resolution matting dataset with 460 images of the transparent or non-salient foreground. The dataset will be released to promote the development of matting research.

\item Experiments on four matting datasets demonstrate that the proposed TransMatting method outperforms the current SOTA methods. And the proposed tri-token could boost both Transformer and convolutional SOTA methods, indicating the great utilization potentiality in image matting.

\end{enumerate}

% I wish you the best of success.

A preliminary conference version of this work appeared in \cite{cai2022transmatting}. On the basics of \cite{cai2022transmatting}, we extend it in threefolds. (1) We extend the network (named as TransMattingV1) proposed in \cite{cai2022transmatting} to TransMattingV2 by introducing our proposed tri-token to both Transformer and convolutional modules. Thus we could boost the performance of any off-the-shelf deep network by replacing the traditional trimap with our tri-token. (2) We further investigate some critical designs of the tri-token, including more introduction positions, initialization methods, and interpretability. (3) By introducing the tri-token to both the CNN local extractor and Transformer blocks, we show the superiority of TransMatting in both composite and real-world images.

\section{Related Works}

 \subsection{Traditional Matting Methods}
Traditional matting methods can be divided into two categories: sampling-based and propagation-based methods. These methods mainly rely on low-level features, like color, location, etc. The sampling-based methods \cite{berman2000method,chuang2001bayesian,wang2007optimized,gastal2010shared,he2011global,shahrian2013improving} first predict the colors of the foreground and background by evaluating the similarity of colors between the known foreground, background and unknown area in samples, and then predict alpha mattes. Various sampling techniques have been investigated, including colour cluster sampling\cite{shahrian2013improving}, edge sampling\cite{he2011global}, ray casting\cite{gastal2010shared}, etc. The propagation-based methods \cite{chen2013knn,lee2011nonlocal,sun2004poisson} propagate the information from the known foreground and background to the unknown area by solving the sparse linear equation system\cite{levin2007closed}, the Poisson equation system\cite{grady2005random}, etc., to obtain the best global optimal alpha. 

\subsection{Deep-Learning Matting Methods}
In recent decades, deep learning technologies have boomed in various fields of computer vision. The same goes for the image matting task. \cite{tang2019learning} combines the sampling and deep neural network to improve the accuracy of alpha matting prediction. 

Based on providing a larger dataset Composition-1k \cite{xu2017deep}, DIM utilizes an encoder-decoder model to directly predict alpha mattes, which effectively improves the accuracy. 
IndexNet~\cite{lu2019indices} proposes a learnable index function to unify upsampling operators for image matting and other three dense prediction tasks.
\cite{sun2021semantic} introduces semantic classification information of the matting region and uses learnable weights and multi-class discriminators to revise the prediction results. \cite{yu2021mask} proposes a general matting framework, which is conducive to obtaining better results under the guidance of different qualities and forms. \cite{liu2021tripartite} further mines the information of the RGB map and trimap and fuses the global information from these maps for obtaining better alpha mattes. 
All of the above methods use trimap as guidance. Some trimap-free methods can predict alpha mattes without using trimap. However, the accuracy of these trimap-free methods still has a big gap compared to that of the trimap-guided ones \cite{chen2018semantic,yang2018active,qiao2020multi,yang2020smart}, indicating that the trimap could help the model to capture information efficiently.

\subsection{Vision Transformer}
The Transformer is firstly proposed in \cite{vaswani2017attention} to model long-range dependencies for machine translation and has demonstrated impressive performance on NLP tasks. Inspired by this, numerous attempts have been made to adapt transformers for vision tasks, and promising results have been shown for vision fields such as image classification, objection detection, semantic segmentation, etc. In particular, ViT \cite{dosovitskiy2020image} divides the input image into patches with a size of 16 $\times$ 16 and feeds the patch sequences to the vanilla Transformer model. To help the training process and improve the performance, DeiT \cite{touvron2021training} proposes a teacher-student strategy, which includes a distillation token for the student to learn from the teacher. Later, Swin \cite{liu2021swin}, PVT \cite{wang2021pyramid}, Crossformer \cite{wang2021crossformer}, and HVT \cite{pan2021scalable} combine the Transformer and pyramidal structure to decrease the number of patches progressively for obtaining multi-scale feature maps. To reduce computing and memory complexity, Swin, HRFormer \cite{yuan2021hrformer}, and CrossFormer apply local-window self-attention in Transformer, which also shows superior or comparable performance compared to the counterpart CNNs. The powerful self-attention mechanism in Transformer shows great advantages over CNN by capturing global attention of the whole image. However, some researchers \cite{li2021localvit} argue that locality and globality are both essential for vision tasks. Therefore, various researchers have tried combining the locality of CNN with the globality of Transformer to improve performance further. LocalViT \cite{li2021localvit} brings depth-wise convolutions to vision transformer to combine self-attention mechanism with locality, and shows great improvement compared to the pure Transformer, like DeiT, PVT, and TNT \cite{han2021transformer}.

\section{Matting Dataset}

According to the transparency of foregrounds, we could divide the images into two types: 1) Transparent Partially (TP): there are significant foreground and uncertainty areas, and the foreground areas can provide information for the prediction of uncertainty areas. For example, when the foreground is human, the opaque and unknown regions are the hair or clothes. 2)  Transparent Totally (TT): there are minor or non-salient foreground areas, and the entire image is semi-transparent or highly transparent. These images include glass, plastic bags, fog, water drops, etc.

As illustrated in Tab.~\ref{tab:dataset}, we select four popular image matting datasets for comparison, including DAPM \cite{shen2016deep}, Composition-1k \cite{xu2017deep}, Distinctions-646 \cite{qiao2020attention}, and AIM-500 \cite{li2021deep}. The DAPM dataset only consists of portraits with no translucent or transparent objects. The Composition-1k dataset contains multiple categories, while most images are portraits and there are only 86 TT-type objects among them. The Distinctions-646 dataset also mainly consists of portraits and has a similar number (79) of TT objects.
The same condition is also in the AIM-500 dataset, which contains only 76 (15.2\%) TT-type images (corresponding to the Salient Transparent/Meticulous type and the Non-Salient type as defined in the original dataset). 

As we can see, the transparent objects in the above datasets only occupy a small portion. This may be because it is much more difficult to label transparent objects than other objects, limiting the progress of transparent objects in the matting field. In this work, we propose the first large-scale dataset targeting various highly transparent objects called \Ourds\ dataset. Our \Ourds\ dataset includes 460 high-quality manually-annotated alpha mattes, where 410 images are for training and 50 for testing. Furthermore, to our best knowledge, the resolution of our \Ourds\ is the highest (the average resolution is up to 3820 $\times$ 3766) among all datasets with high transparent objects. We believe this new matting dataset will greatly advance the matting research on objects with massive transparent areas. 

%\IEEEpubidadjcol

\begin{table}
\centering
\caption{Comparison between different public matting datasets. }
\label{tab:dataset}
\begin{tabular}{cccc} \toprule
Image Matting Dataset      & Total Num & TT Num & Resolution \\ \midrule
DAPM \cite{shen2016deep}           & 2000  & 0       & 800$\times$600    \\
Composition-1k \cite{xu2017deep}  & 481   & 86      & 1297$\times$1082  \\
Distinction-646 \cite{qiao2020attention} & 646   & 79      & 1727$\times$1565  \\
AIM-500 \cite{li2021deep}         & 500   & 76      & 1260$\times$1397  \\
\Ourds~(Ours)                       & 460   & 460     & 3820$\times$3766  \\ \bottomrule
\end{tabular}
\end{table}

\section{Methodology}

\begin{figure*}[t]
\centering
\includegraphics[width=0.99\textwidth]{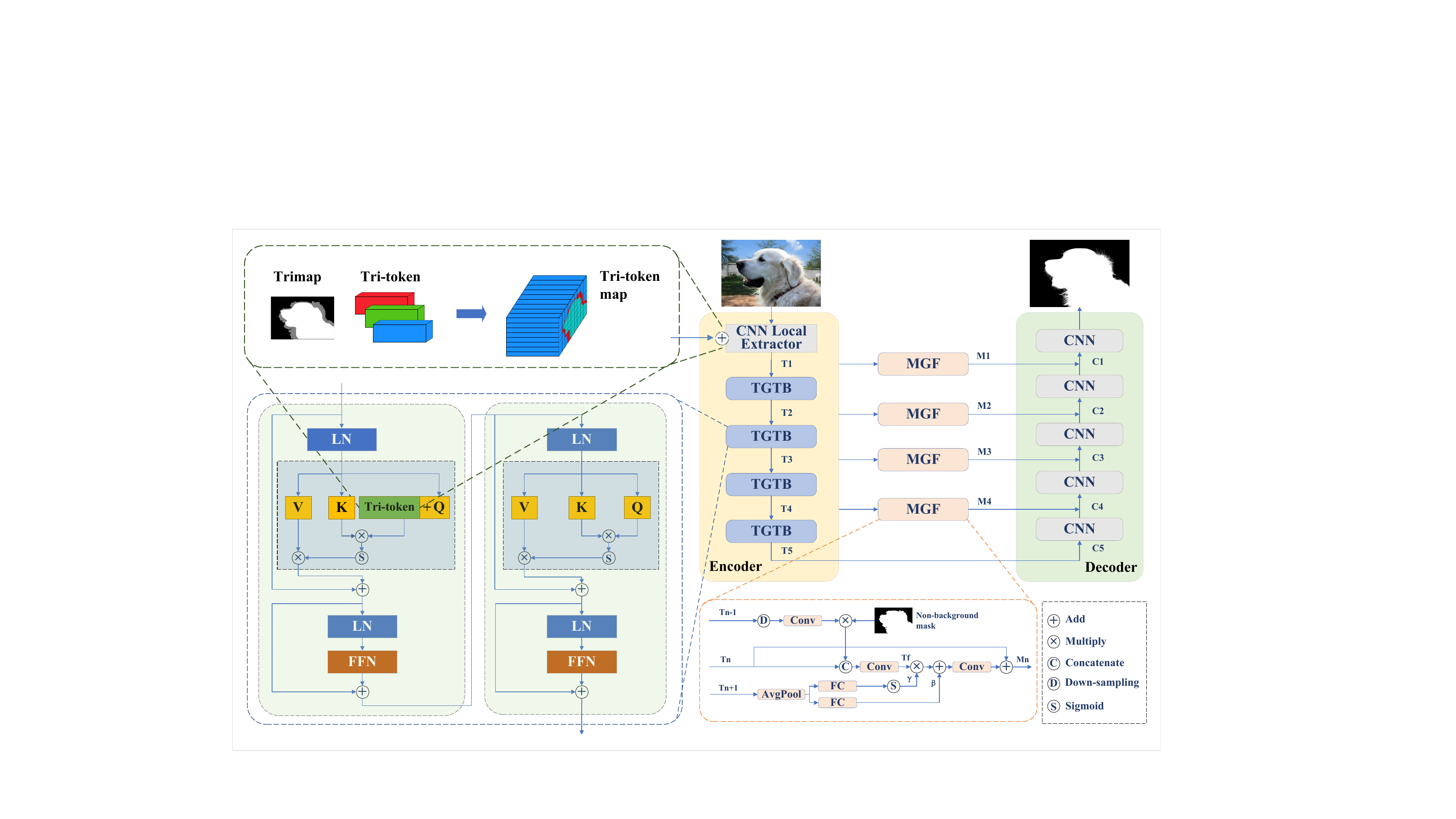}
\caption{The structure of our TransMatting. Unlike vanilla matting methods that concatenate the trimap with the image, we feed the RGB image into the CNN local extractor and introduce tri-token directly to extracted deep features. We design two ways to introduce tri-token to both convolutional and Transformer models. To better model transparent objects with few known foregrounds, the TGTB module is proposed to extract further long-range features. And the MGF module is responded to fusing multi-level features with global guidance.  }
\label{fig:model}
\end{figure*}

\subsection{Baseline Structure}

To extract both local and global features, we combine CNN and the Transformer model as our baseline encoder, as illustrated in Fig.~\ref{fig:model}. 
Specifically, based on the ResNet34-UNet \cite{yu2021mask,li2020natural}, we only use the first two stage of the ResNet34 as our CNN local extractor. And then, 
% The second part consists of 
a stack of Swin Transformer blocks \cite{liu2021swin} are adopted as the Transformer encoder. Other components are the same as the original ResNet34-UNet. It is worth noting that this Transformer-based baseline structure achieves a comparable performance with current matting methods. And we will introduce our main contribution as follows.

\subsection{Tri-token}
Almost all SOTA methods \cite{liu2021tripartite,yu2021mask,li2020natural,xu2017deep,yu2020high,sun2021semantic,cai2019disentangled} use trimap as a guide and directly concatenate the RGB image with the trimap as the model's input. However, the modalities of the RGB image and trimap are quite different. The RGB image scales from 0 to 255 and shows fine low-level features like texture, color similarity, etc. The trimap includes three values containing high-level semantic information, like shape, location, etc., \cite{liu2021tripartite}. It is not suitable to use one extract to extract features from these two modalities.
Furthermore, 
% since the vision tasks' mainstream paradigm for the input is RGB images, 
most deep models utilize pre-trained weights on the classification task (with three channels in the input) for backbone initialization, 
the concatenation of the RGB image and the trimap as input (four channels) will break the pre-trained weights and affect the performance.
Thus, their direct concatenation is not the most efficient way to extract features from the trimap.

To the best of our knowledge, we are the first to attempt to explore different harmonising manners of the RGB image and trimap rather than simply concatenating them. 

Inspired by the \texttt{[cls]} token in Vision Transformer, we design a new tri-token (shown in Fig.~\ref{fig:model}) manner, aiming to introduce the high-level semantic information directly into the deep features to replace the inefficient concatenation method. 

Given a vanilla $Trimap \in \mathbb{R}^{H \times W}$, we generate three \textbf{learnable} tokens (denoted as $Token_i$, $i=\{0,1,2\}$) with different initialization to represent the known foreground, known background, and unknown areas, respectively. Every token is a 1D vector, that is, $Token_i \in \mathbb{R}^{C}$.
Then we replace every pixel in the trimap with the corresponding tri-token values to generate the tri-token map, formulated as:
\begin{equation}
\label{eq:tri-token-map}
    Trimap[Trimap==i] = Token_i,   i=\{0,1,2\}
\end{equation}

In this manner, the trimap is converted to a more flexible manner. With an additional channel dim, the tri-token could easily introduce semantic and location information to high-level abstract feature maps. Besides, the learnable tri-token have the ability to learn to align trimap spaces and image representation spaces. The following will introduce how to utilize tri-token to guide both Transformer and convolutional models.

\subsection{Tri-token Guided Image Matting Network}

\subsubsection{Tri-token Guided Transformer Block}  
Due to the large proportion of uncertainty areas in images of transparent objects, global connectivity is much more important to mine features from far known patches. CNN does not have a big enough receptive field to cover the whole foreground object \cite{liu2021long}, which leads to poor estimation of pixels outside receptive fields. In contrast, the receptive field of the Transformer could extend to the whole image with the self-attention mechanism. Thus we adopt the Transformer model here to extract long-range features of transparent objects.

The Transformer model consists of multi-head self-attention (MHSA) and Multi-layer Perception (MLP) modules. The self-attention mechanism can be thought of as a mapping between a query and a collection of key-value pairs. The output is a weighted sum of the values, and the weights are assigned by the compatibility function between the query and the relevant key. This can be implemented by Scaled Dot-Product Attention \cite{vaswani2017attention}, in which a $softmax$ function is used to activate the dot products of query and all keys for obtaining the weights. MHSA means that more than one self-attention is performed in parallel.

Like \cite{liu2021swin,wang2021crossformer,yuan2021hrformer}, we use non-overlapping windows whose size is $M \times M$ to divide the feature maps. The MHSA is performed within each window. The formulations of vanilla attention and our tri-token attention in a specific window are shown as follows:
\begin{equation}
    Attention(Q, K, V) = \mathcal{S}(QK^T / \sqrt{d})V
\end{equation}

\begin{small}
\begin{equation}
    Tri\mbox{-}token\ Attention(Q, K, V) = \mathcal{S}((Q + Tri\mbox{-}token)K^T / \sqrt{d})V
\end{equation}
\end{small}

where $Q, K, V \in \mathbb{R}^{M^2 \times d}$ represent the query, key, and value in the attention mechanism, respectively. $d$ is the query/key dimension and $\mathcal{S}$ denotes the $softmax$ function. In the Tri-token Attention formulation, $Q$, $K$, and $V$ are the same as that in the standard self-attention. The $Tri\mbox{-}token$ is our proposed learnable \emph{trimap} that adds to the query for forming a new tri-token guided query. In this way, our tri-token attention mechanism can selectively aggregate contexts and evaluate which region should be paid more attention to with the guidance of our learnable tri-tokens.

In this way, we combine the self-attention and the tri-token to focus on more valuable regions by considering the relationship among known foreground, known background and uncertainty areas and finally achieve the best performance. We replace the vanilla Transformer block with our Tri-token Guided Transformer Block (TGTB) every five blocks in the Transformer encoder.
% 

%"mystyle" code listing set
\lstset{style=mystyle}

\begin{algorithm}[t]
\caption{Pseudo codes to convert traditional trimap to our proposed tri-token in PyTorch style.}\label{code}
\begin{lstlisting}[language=Python]
trimap = cv2.imread('trimap.png', 0)  # [H, W]
H, W = trimap.shape

# convert to tri-token
dim = 64
foreground_token = nn.Parameter(torch.zeros(1, 1, dim))
uncertain_token = nn.Parameter(torch.zeros(1, 1, dim) + 1)
background_token = nn.Parameter(torch.zeros(1, 1, dim) + 2)
# build a tri-token map for convenience
tri_token_map = torch.zeros(H, W, dim)
tri_token_map[trimap == 0] = background_token
tri_token_map[trimap == 128] = uncertain_token
tri_token_map[trimap == 255] = foreground_token
\end{lstlisting}
\end{algorithm}

\subsubsection{Tri-token Guided Convolutional Network}
As convolutional networks are still commonly used for image matting, it is quite meaningful to extend our tri-token to boost current CNN-based matting methods. Different from the Transformer model, which is based on the self-attention mechanism, convolutional networks mainly consist of a stack of convolutional layers. There are no ready-made attention modules in it to introduce tri-token information. To make our modification general and simple, we employ the additive operation (formulated in Eq.~\ref{eq:conv-tritoken}) for convolutional models. Specifically, after some convolution layers, a feature map is extracted from the input image. Every pixel vector in the feature map represents the semantic features of a small input image patch. According to the trimap, we could add the corresponding tri-token to the pixel vector to introduce high-level tri-token information into feature maps. In this way, the tri-token can guide the network to distinguish foreground, background, and uncertain areas to pay more attention to where it needs to focus. The $Tri\mbox{-}token\mbox{-}map$ in Eq.~\ref{eq:conv-tritoken} can be generated efficiently with Alg.~\ref{code}.

% \begin{equation}
%     x_1 = Conv_2(Conv_1(x))
% \end{equation}

\begin{equation}
\label{eq:conv-tritoken}
    x = Conv_2(Conv_1(x) + Tri\mbox{-}token\mbox{-}map) 
\end{equation}

\subsection{Multi-scale Global-guided Fusion Module}

In the multi-scale feature pyramid structure, in-depth features contain more global information, while shallow features have rich local information like texture, color similarity, etc. Fusing these features is vital for accurately predicting alpha mattes for high transparent objects \cite{qiao2020multi}.
Although the direct sum operation can realize feature fusion, the details in the shallow features may attenuate the impact of the advanced semantics, resulting in some subtle regions missing \cite{qiao2020multi}. To address this issue, we propose a Multi-scale Global-guided Fusion (MGF) module in the decoder process (see Fig.~\ref{fig:model} for details), with both the non-background information and the advanced semantic features as guidance, to fuse the high-level semantic information and the lower ones effectively.

Specifically, we denote three adjacent features from shallow to deep as $T_{n-1}$, $T_{n}$, and $T_{n+1}$. The $T_{n-1}$ is first down-sampled, then the Hadamard product is employed between the non-background mask and $T_{n-1}$ to extract the low-level features of non-background, which helps to reduce the impact of complex background influence. This can guide the network to pay more attention to the foreground and unknown areas. After that, the $T_{n-1}$ is concatenated with $T_{n}$, and a convolution layer is performed to align the channel of fused features. We mark this feature as $T_{f}$. 

For the $T_{n+1}$, we first perform a global average pooling to generate channel-wise statistics and then use two fully connected (FC) layers to squeeze channels. As shown in Fig.~\ref{fig:model}, features output from the two FC layers are denoted as $\gamma$ and $\beta$, separately.To fully capture channel-wise dependencies, we add a sigmoid function to activate $\gamma$ and perform broadcast multiplication with $T_{f}$ for channel re-weighting. After that, broadcast addition is performed between the channel-weighted feature and $\beta$. A convolution layer is used to fuse information from different groups. Notably, a skip connection from $T_{n}$ is employed for obtaining the final fused features of MGF.

In short, considering that fusing low-level features directly may cause a negative impact on the advanced semantics \cite{qiao2020multi}, two techniques are proposed here. Firstly, the non-background mask is introduced into the fusion process to filter out the complex background information and further help to concentrate more attention on the foreground and unknown areas. Secondly, the global channel-wise attention from higher-level features is used for re-weighting and enhancing the important information in the fused features.

\subsection{Loss Function}
Following \cite{yu2021mask}, we use three kinds of losses, including alpha loss ($\mathcal{L}_\alpha$), compositing loss \cite{xu2017deep} ($\mathcal{L}_{comp}$), and Laplacian loss \cite{hou2019context} ($\mathcal{L}_{lap}$). As formulated below, the weights are set as 0.4, 1.2, and 0.16, respectively.

\begin{equation}
    \mathcal{L}_{final} = 0.4*\mathcal{L}_{alpha} + 1.2*\mathcal{L}_{comp} + 0.16*\mathcal{L}_{lap}
\end{equation}

\section{Experiments}

\begin{figure*}
\centering
\includegraphics[width=0.79\textwidth]{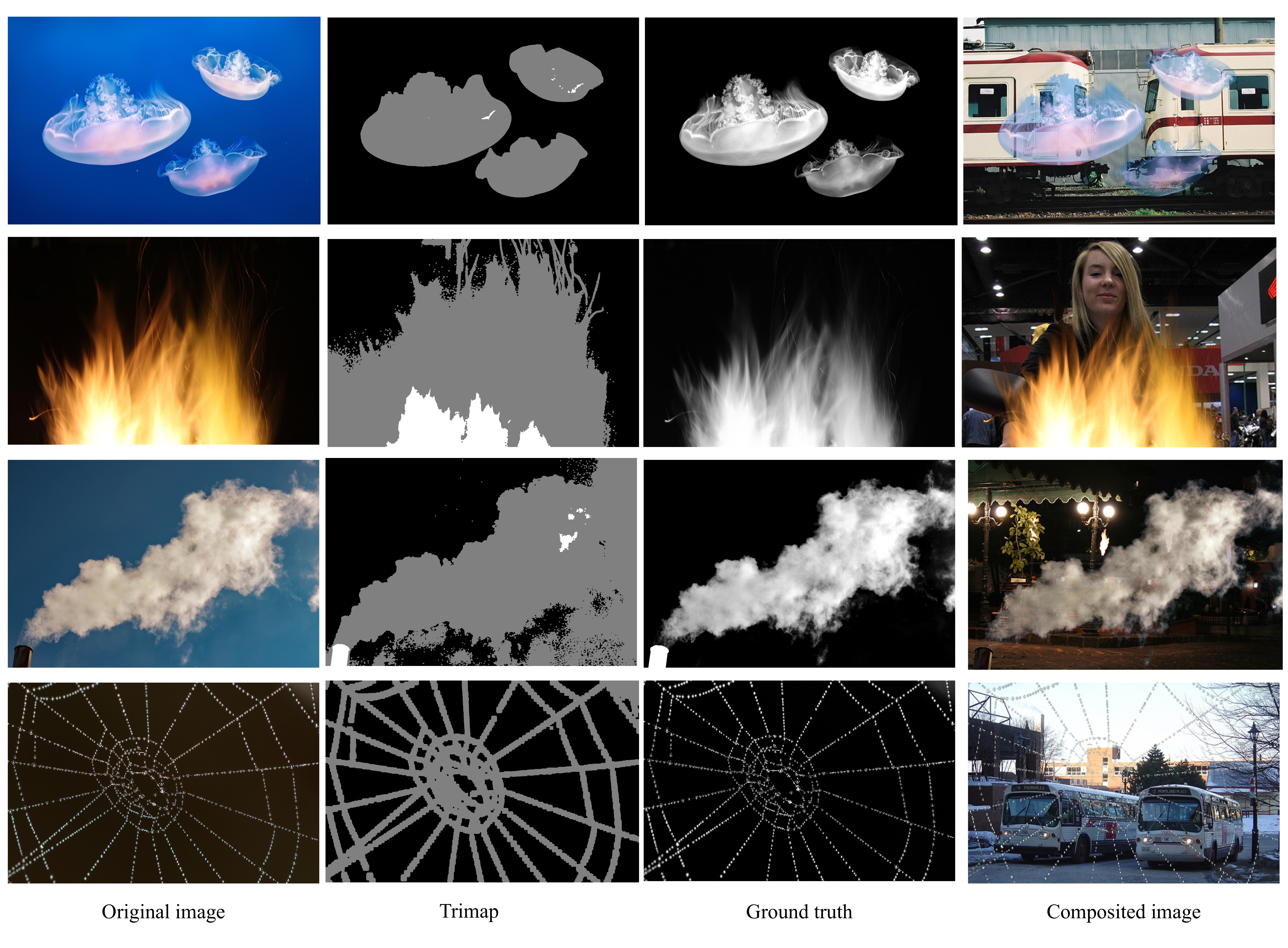}
\caption{Samples from our Transparent-460 dataset. 
The foreground objects are highly transparent and occupy a large area of the image, which is very challenging for current image matting methods.
All images are high-resolution. Thus, it is better to zoom in to find subtle details.
}
\label{fig:Transparent460}
\end{figure*}

\subsection{Dataset}
% We evaluate our proposed method on Composition-1k, Distinction-646, and our Global Transparency Matting(GTM) dataset.

\textbf{Composition-1k} \cite{xu2017deep} contains 493 and 50 unique foreground objects as training and test sets, respectively. Every foreground object is composited with 100 background images from COCO \cite{lin2014microsoft} and 20 from Pascal VOC \cite{everingham2010pascal}. As a result, there are 49,300 images for training and 1,000 images for testing.

\textbf{Distinction-646} \cite{qiao2020attention} comprises 646 distinct foreground objects. Similar to the Composition-1k, 50 objects are divided as the test set. Following the same composition rule, there are 59,600 and 1000 images for training and testing, respectively.

\textbf{AIM-500} \cite{li2021deep} is a real-world benchmark, including many types of objects, like humans, glass, grids, etc. It contains 500 natural images for testing the performance of real-world matting methods.

\textbf{Our Transparent-460} mainly consists of transparent and non-salient objects as the foreground, like water drops, jellyfish, plastic bags, glass, crystals, etc. We collect 460 high-resolution images and carefully annotate them with Photoshop. 
Considering the transparent objects are very meticulous,  we keep the origin resolution of all collected images up to 3820 $\times$ 3766 on average. To our best knowledge, this is the first transparent object matting dataset in such a high resolution. Some samples are illustrated in Fig.~\ref{fig:Transparent460}.

\subsection{Evaluation Metrics}
Following \cite{hou2019context,cai2019disentangled,lu2019indices,liu2021tripartite}, we use four metrics for evaluation, including the Sum of Absolute Differences (SAD), Mean Squared Error (MSE), Gradient error (Grad.) and Connectivity error (Conn.). It is notable that the unit of MSE value is set to 1e-3 for easy reading.

%\IEEEpubidadjcol

\subsection{Implementation Details}
We use the PyTorch \cite{paszke2019pytorch} to implement our proposed method. All the experiments are trained for 200,000 iterations. We initialize our network with ImageNet \cite{deng2009imagenet} pre-trained weights and initialize the background, uncertainty, and foreground token in the tri-token with values of 0, 1, and 2, respectively. The ablation experiments in Tab.~\ref{tab:ab:main}, \ref{tab:initializastion}, \ref{tab:position}, \ref{tab:tri-tokens-positions-cnn} are done with 2 NVIDIA Tesla V100 GPUs with a batch size of 32. Moreover, to compare our method with the existing SOTA methods, we use a batch size of 64 with 4 NVIDIA Tesla V100 GPUs to train our proposed method in Tab.~\ref{tab:sota:AIM}, \ref{tab:sota:Distinctions}, \ref{tab:sota:ours}. The Adam optimizer is utilized, and the initial learning rate is set to 1e-4 with the same learning rate decay strategy in \cite{yu2021mask,loshchilov2016sgdr}. For a fair comparison, we follow the same data augmentation as \cite{li2020natural}, like random crop, rotation, scaling, shearing, etc. Moreover, trimaps for training are generated using dilation and erosion ways on alpha images by random kernel sizes from 1 to 30. Finally, we crop 512×512 patches on the center of the unknown area of alpha and composition with backgrounds from COCO. The same settings are applied to Composition-1k, Distinction-646, and our collected Transparent-460.

%\IEEEpubidadjcol

\subsection{Ablation Study}

To evaluate different designs of our new proposed tri-token and MGF module,  we designed some ablation studies on the Composition-1k test set.

\begin{table}[t]
\setlength\tabcolsep{5pt}
\centering
\caption{The effectiveness of our proposed tri-token strategy and MGF module on the Composition-1k test set.}
\label{tab:ab:main}
\begin{tabular}{ccc|cccc}
\toprule
\multicolumn{2}{c}{Tri-token}   &  \multirow{2}{*}{MGF}       & \multirow{2}{*}{SAD $\downarrow$}   & \multirow{2}{*}{MSE $\downarrow$}  & \multirow{2}{*}{Grad. $\downarrow$}  & \multirow{2}{*}{Conn.$\downarrow$}  \\
% \multicolumn{2}{c}{Tri-token}   & MGF        & SAD $\downarrow$   & MSE $\downarrow$  & Grad. $\downarrow$  & Conn.$\downarrow$  \\
\cline{1-2}
CNN & Transformer & \\ \midrule
 &      &       & 29.14     & 6.34    & 12.06     & 25.21     \\
   & $\checkmark$ & & 27.45     & 5.66   & 11.77     & 24.30\\
    
 $\checkmark$  & $\checkmark$ &  & 26.40	& 5.22	& 11.03	& 22.47 \\
    & & $\checkmark$  & 27.21     & 5.57    & 11.23     & 23.25     \\ 
 & $\checkmark$ & $\checkmark$ & 26.83 & 5.22 & 10.62 & 22.14 \\
$\checkmark$  &  & $\checkmark$ & 26.43	&5.24	&10.97	&22.41 \\

 $\checkmark$ & $\checkmark$ & $\checkmark$ & \textbf{26.36} & \textbf{5.08} & \textbf{10.24} & \textbf{21.68} \\
 \bottomrule
\end{tabular}
\end{table}

\textbf{Evaluate the effectiveness of our proposed modules.} The quantitative results under the SAD, MSE, Grad., and Conn. with and without our proposed tri-token and MGF module are illustrated in Tab.~\ref{tab:ab:main}. As we can see, with the tri-token introduced in Transformer (TGTB), the four metrics decrease from 29.14, 6.34, 12.06, and 25.21 to 27.45, 5.66, 11.77, and 24.30, respectively. Combined with introducing the tri-token in the CNN local extractor, the prediction error could further decrease to 26.40, 5.22, 11.03, and 22.47, respectively. It indicates that,  for both the Transformer and convolutional networks, our redesigned tri-token is more effective in helping the model to distill the advanced semantics and pay attention to critical areas than traditional concatenating to the input image. Compared with the baseline model (the first row), the addition of the MGF module could solely achieve similar performance, which decreases by 1.93, 0.77, 0.83, and 1.96 on four metrics. This indicates that our proposed multi-scale feature fusion strategy can also help the decoder to harmonize local and global features to guide information integration. On the basis of the MGF module, whether adding the tri-token to the convolutional or Transformer module, our model can achieve better performance. When combined with the tri-token on two networks and the MGF module, the model achieves the best performance of 26.36, 5.08, 10.24, 21.68 in terms of SAD, MSE, Grad., and Conn., which improves 2.78, 1.26, 1.82, and 3.53 compared with the baseline strategy, showing the effectiveness of our proposed tri-token and the MGF module.

\begin{table}[t]
\setlength\tabcolsep{5pt}
    \centering
    \caption{Quantitative results on the Composition-1k test set with different initialization methods of tri-token.}
    \label{tab:initializastion}
    \begin{tabular}{cccccc}
    \toprule
    Method  & Init Method     & SAD $\downarrow$   & MSE $\downarrow$  & Grad. $\downarrow$  & Conn. $\downarrow$  \\ \midrule
\multirow{3}{*}{MGMatting}  &  (-1, 0, 1)    & 28.56	&5.53	&11.61	&24.98 \\
  & (0, 1, 2)     & \textbf{28.34}	&\textbf{5.45}	&\textbf{10.96}	&\textbf{24.62} \\
  & random     & \textbf{28.34}	&5.51	&11.12	&24.70 \\ \hline
\multirow{3}{*}{TransMattingV2} & (-1, 0, 1)    & 26.81	& 5.16	& 11.24	& 23.11  \\
  & (0, 1, 2)     & \textbf{26.36}	&\textbf{5.08}	&\textbf{10.24}	&\textbf{21.68}  \\
 & random     & 26.38	&5.09	&10.68	&22.50 \\
\bottomrule
\hline
\end{tabular}
\end{table}

\textbf{Determine different initialization methods for tri-token.} 
Different from the vanilla trimap, our proposed tri-token is learnable and could update itself as the model training. 
Considering that initialization is essential for deep learning methods, we further explore different approaches to initialize our tri-token: pre-defined values for every token or totally random. 
For the pre-defined manner, we initialize background, uncertainty, and foreground with values $a$, $b$, and $c$, respectively, denoted as ($a$, $b$, $c$). We conduct experiments on three initialization approaches: (-1, 0, 1), (0, 1, 2), and random initialization. The results are illustrated in Tab.~\ref{tab:initializastion}. As we can see, the proposed tri-token is robust with different initialization methods.
Among these methods, we can observe that the (0, 1, 2) initialization performs best for both CNN-based MGMatting and our Transformer-based TransMatting. But the improvement is limited, indicating that the initialization method is not critical for our tri-token updating.

\textbf{Determine where to introduce the tri-token in our TransMatting.}
The tri-token is proposed to introduce high-level semantic and position information to the deep model. As we know, the shadow features in deep models have a big resolution which reserves more position information. In contrast, the deep layer could extract more abstract semantic features, which is suitable for mutual learning with the tri-token.
To determine which position is more suitable to introduce the tri-token, we conduct experiments with different positions of the Transformer encoder in our TransMatting model.
There are four stages in the Transformer encoder. We adopt the stage number  $i$ ($i\in\{1,2,3,4\}$) to represent the insertion position. For example, position 3 indicates that the tri-token is introduced at the beginning of stage 3. The experiment results on Composition-1k are illustrated in Tab.~\ref{tab:position}. By comparing the performance of different introduction positions for tri-token, we conclude that introducing tri-token in multiple positions achieves the best performance.

\begin{table}[t]
    \centering
    \caption{Quantitative results on the Composition-1k test set with different positions to introduce tri-token into our TransMattingV1. 
    }
    \label{tab:position}
    \begin{tabular}{@{}ccccc@{}}
    \toprule
    Position        & SAD $\downarrow$   & MSE $\downarrow$  & Grad. $\downarrow$  & Conn. $\downarrow$  \\ \midrule
1     & 31.68 & 7.24 & 14.20 & 27.42  \\
2     & 29.32 & 6.66 & 13.42 & 25.16 \\
3     & 27.25 & 5.34 & 10.89 & \textbf{22.12} \\
4     & 29.50 & 6.20 & 13.18 & 25.23  \\
3, 4  &  27.12 & \textbf{5.20} & 11.07 & 22.59 \\
% 5     & 26.43	&5.24	&10.97	&22.41 \\
1,2,3,4       & \textbf{26.83} & 5.22 & \textbf{10.62} & 22.14 \\ 
% 1,2,3,4,5    & \textbf{26.36}	&\textbf{5.08}	&\textbf{10.24}	&\textbf{21.68} \\
\bottomrule

\hline
\end{tabular}
\end{table}

\begin{table}[t]
\setlength\tabcolsep{5pt}
\centering
\caption{Quantitative results on the Composition-1k test set with different positions to introduce the tri-tokens into convolutional networks on SOTA CNN methods.
}
\label{tab:tri-tokens-positions-cnn}
\begin{tabular}{cccccc} \toprule
Methods & Position      & SAD $\downarrow$ & MSE $\downarrow$ & Grad. $\downarrow$ & Conn. $\downarrow$ \\ \midrule
\multirow{6}{*}{GCAMatting~\cite{li2020natural}} & - & 35.3  & 9.1 & 16.9 & 32.5  \\
 &  1   & 33.71	&7.75	&15.80	&30.28 \\
 &  2  & 33.79	&8.09	&15.51	&30.16 \\
 &  3  & 32.72	&7.48	&14.05	&29.19 \\
 &  4  & \textbf{31.32}	&\textbf{7.02}	&\textbf{13.24}	&\textbf{27.67} \\
 &  3,4 & 32.41 & 7.45 & 14.62 & 29.08\\ 
 &  1,2,3,4   & 32.38	& 7.52	&14.92	&28.77\\ 
\hline
\multirow{6}{*}{MGMatting~\cite{yu2021mask}} & - & 31.5  & 6.8  & 13.5 & 27.3 \\
 &  1    & 28.34	&5.45	&11.08	&24.83 \\
 &  2   & 28.85	&5.52	&11.43	&24.90 \\
 &  3   & \textbf{28.31}	&5.52	&11.37	&24.73 \\
 &  4   & 28.32	&\textbf{5.41}	&\textbf{10.96}	&\textbf{24.62} \\
 &  3,4  & 28.82	& 5.61	& 11.39	&25.12 \\
 & 1,2,3,4  & 28.59	&5.59	&11.37	&24.99\\ 
\bottomrule

\hline
\end{tabular}
\end{table}

\textbf{Determine where to introduce tri-token in convolutional networks.}
As the introduction manner of the tri-token in convolutional models differs from the Transformer, we conduct experiments on Composition-1k to find the best introduction position for convolutional networks. Since our TransMatting only has a few layers in the  CNN local extractor, we conduct experiments on two convolutional SOTA methods: GCAMatting and MGMatting. The results are illustrated in Tab.~\ref{tab:tri-tokens-positions-cnn}. The position ``-" indicates no tri-token introduced, and the trimap is concatenated to the image. As we can see, replacing trimap with our tri-token achieves significant improvements on all metrics with any position, indicating that our tri-token is a highly flexible and efficient design. We can also find that the performance gets better as the introduction position goes deeper. The best performance is achieved when tri-token is introduced to stage four: outperform the trimap strategy with about 3 - 4 improvements on SAD. This indicates that the information in trimap is really high-level and more suitable for interacting with deep features in convolutional models. 

However, unlike the Transformer model, introducing the tri-token to multiple positions decreases the performance of GCAMatting. The SAD and MSE increase from 31.32 and 7.02 to 32.41 and 7.45, respectively, when introducing the tri-token to both stage 3 and stage 4. The performance further decreases to 32.38 and 7.52 when introduced to all four stages. The same trends are observed in MGMAtting. We guess this is caused by the different introduction manner between convolutional models and Transformer ones. The straightforward addition operation makes it hard for the tri-token to simultaneously suit different features from different stages.

\subsection{Comparison with Prior Works}
To evaluate the performance of our proposed method, we compare it with current SOTA image matting models on three synthetic image matting datasets (Composition-1k, Distinction-646, and our collected Transparent-460) and one real-world dataset (AIM-500). 

\subsubsection{Experiments on Synthetic Datasets}

\begin{table}
\caption{The quantitative results on Composition-1k test set.  $^{\dagger}$ denotes results with test-time augmentation. 
% $^{\ast}$ denotes methods with tri-token introduced in the convolutional networks.
Following IndexNet, the matting refinement stage is not applied in DIM$^{\ast}$ for simplicity and fair comparison.
% . 
} 
\label{tab:sota:AIM}
\centering
\begin{tabular}{l|cccc}
\toprule
Methods & SAD $\downarrow$& MSE $\downarrow$ & Grad. $\downarrow$ & Conn. $\downarrow$ \\ \midrule
AlphaGAN \cite{lutz2018alphagan}      & 52.4  & 30  & 38   & 53    \\
DIM \cite{xu2017deep}           & 50.4  & 14  & 31.0 & 50.8  \\
IndexNet \cite{lu2019indices}      & 45.8  & 13  & 25.9 & 43.7  \\
AdaMatting \cite{cai2019disentangled}    & 41.7  & 10  & 16.8 & -     \\
ContextNet \cite{hou2019context}    & 35.8  & 8.2 & 17.3 & 33.2  \\
TIMI-Net \cite{liu2021tripartite}      & 29.08 & 6.0 & 12.9 & 27.29 \\
FBAMatting \cite{forte2020f} $^{\dagger}$     & 25.8  & 5.2 & 10.6 & 20.8  \\ \hline
DIM$^{\ast}$ \cite{xu2017deep}           & 54.6	&17	&36.7	&55.3  \\
DIM$^{\ast}$ + Tri-token          & \cellcolor{lightgray!50}\textbf{53.3}	&\cellcolor{lightgray!50}\textbf{16.9}	&\cellcolor{lightgray!50}\textbf{33.54}	&\cellcolor{lightgray!50}\textbf{53.64}  \\ \hline
GCAMatting \cite{li2020natural}    & 35.3  & 9.1 & 16.9 & 32.5  \\
GCAMatting + Tri-token    & \cellcolor{lightgray!50}\textbf{31.32}	&\cellcolor{lightgray!50}\textbf{7.02}	&\cellcolor{lightgray!50}\textbf{13.24}	&\cellcolor{lightgray!50}\textbf{27.67} \\ \hline
MGMatting \cite{yu2021mask}    & 31.5  & 6.8  & 13.5 & 27.3 \\ 
MGMatting + Tri-token  & \cellcolor{lightgray!50}\textbf{28.32}	&\cellcolor{lightgray!50}\textbf{5.41}	&\cellcolor{lightgray!50}\textbf{10.96}	&\cellcolor{lightgray!50}\textbf{24.62} \\ 
\hline
TransMattingV1 & \cellcolor{lightgray!50}24.96 &\cellcolor{lightgray!50} 4.58 & \cellcolor{lightgray!50}9.72 &\cellcolor{lightgray!50} 20.16  \\
TransMattingV2 & \cellcolor{lightgray!50}\textbf{23.82}&\cellcolor{lightgray!50} \textbf{4.37} &\cellcolor{lightgray!50} \textbf{9.43} & \cellcolor{lightgray!50}\textbf{18.86}  \\
\bottomrule
\end{tabular}
\end{table}

\begin{figure*}[!tb]
\centering
\includegraphics[width=0.9\textwidth]{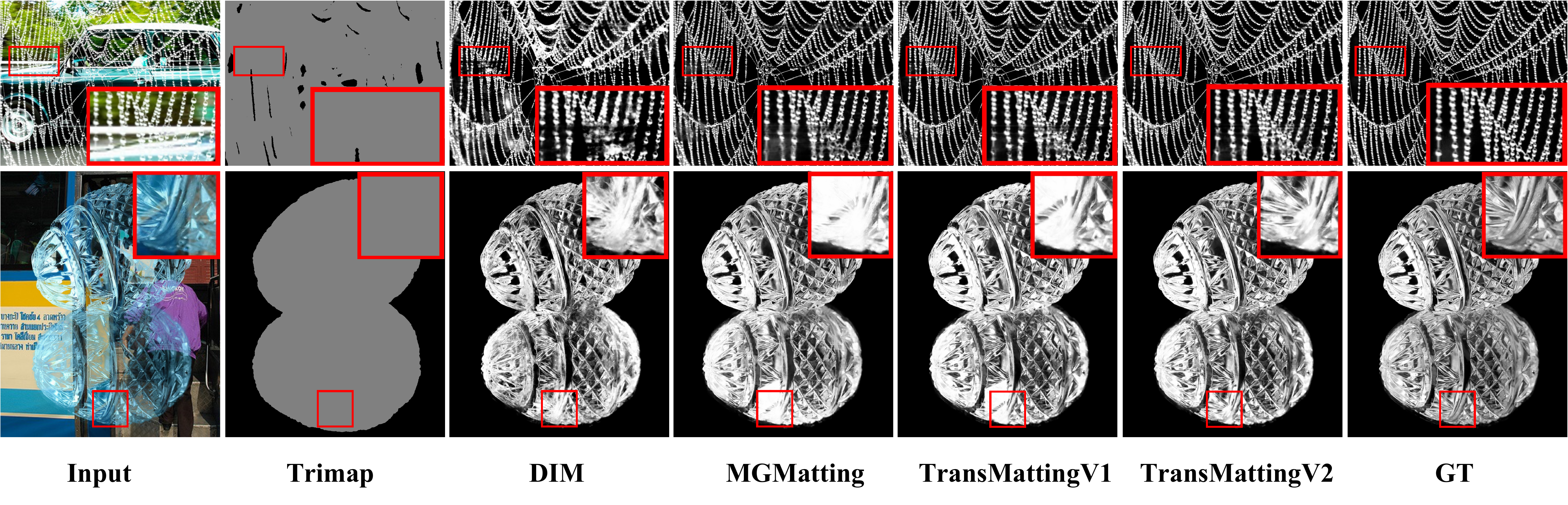}
\caption{Visual comparison of our TransMatting against SOTA methods on the Composition-1k test set.}
\label{fig:composition-1k-test-set}
\end{figure*}

\begin{figure*}[!tb]
\centering
\includegraphics[width=0.9\textwidth]{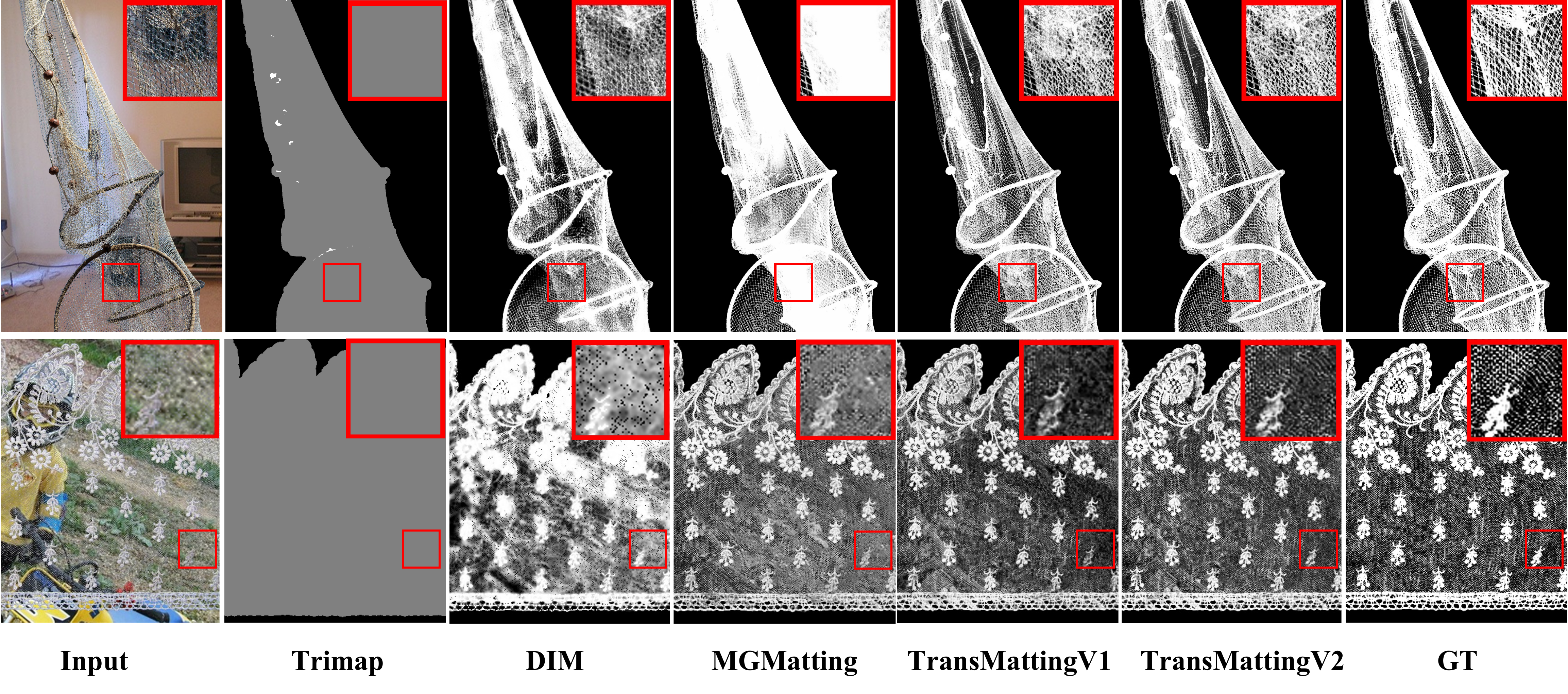}
\caption{Visual comparison of our TransMatting against SOTA methods on the Distinction-646 test set.}
\label{fig:Distinction-646-test-set}
\end{figure*}

\textbf{Testing on Composition-1k.}
We show some quantitative and qualitative results on Tab.~\ref{tab:sota:AIM} and Fig.~\ref{fig:composition-1k-test-set}, respectively. Without any test time augmentations, our proposed TransMattingV1 model outperforms other SOTA methods on all four evaluation metrics by only training on the Composition-1k training set. As illustrated in Tab.~\ref{tab:sota:AIM}, our model decreases the MSE and Grad. heavily: from 5.2, 10.6 to 4.58 and 9.72, respectively, indicating the effectiveness of our TransMattingV1. By further introducing the tri-token to the CNN local extractor, our TransMattingV2 performs better than TransMattingV1, especially on SAD and Conn., improving by 1.14 and 1.3, respectively. Fig.~\ref{fig:composition-1k-test-set} visualizes the qualitative comparisons of two cases on the Composition-1k test set. 
As we can see in the first row, the color of the dewdrop is similar to the background, for which previous methods are all mistaken. In contrast, thanks to the large receptive field and tri-token guided design, our proposed model learns image context and object structure effectively, contributing to sophisticated performance in such a challenging region. Besides, our proposed model can also accurately estimate the TT-type objects whose background is complex. For example, in the second row of Fig.~\ref{fig:composition-1k-test-set}, the background is so complicated that the previous methods are hard to describe the details. It is worth noting that our TransMattingV2 achieves more high-quality alpha mattes than TransMattingV1, indicating that our tri-token helps to distinguish the foreground and background in deep features.

To evaluate the universality of our proposed tri-token, we introduce it to three current SOTA methods: DIM~\cite{xu2017deep}, GCAMatting~\cite{li2020natural}, and MGMatting~\cite{yu2021mask}. We follow~\cite{lu2019indices} to drop the matting refinement stage in DIM for simplicity and fair comparison. The results are illustrated in Tab.~\ref{tab:sota:AIM}. As we can see, our proposed tri-token boosts the performance of all three SOTA methods with a big margin. Especially for recent proposed GCAMatting and MGMAtting, by replacing the traditional concatenation with our proposed tri-token, the SAD is decreased by 3.98 and 3.18, respectively. This proves the effectiveness of the proposed tri-token strategy in existing SOTA methods and indicates a great potential to replace the concatenation strategy with the proposed tri-token in a wide range of image matting methods. 

As illustrated in Tab.~\ref{tab:TTTP}, to further verify our method in boosting the performance of TT-type objects, we analyze two types of objects separately. We can find that our TransMattingV1 achieves 42.4\% and  23.7\% improvements in terms of MSE and SAD on TT-type compared with MGMatting, which is the key factor in improving overall performance. Notably, our TransMattingV2 further improves MSE and SAD by 7.3\% and 6.5\% on TT-type compared to the TransMattingV1. In addition, our model can also achieve a good improvement on TP-type, although not as significant as TT-type. Thus, we can conclude that our proposed method can model TT-type objects with long-range features, and our tri-token design can assist our method in better understanding the semantic information in the trimap and integrating it into deep features as well as long-range features to further improve performance.

\textbf{Testing on Distinction-646.}
Distinction-646 comprises of more distinct foreground images than Composition-1k.
Tab.~\ref{tab:sota:Distinctions} compares the performance of our TransMatting model with other SOTA methods on Distinction-646. For a fair comparison, we follow the whole inference protocol in \cite{qiao2020attention,yu2021mask} to calculate the metrics based on the whole image. Without any additional tuning, our TransMattingV1 outperforms all the SOTA methods. Similarly, our TransMattingV2 could further improve from introducing tri-token in convolutional layers.
We also illustrate some visual comparisons in Fig.~\ref{fig:Distinction-646-test-set}, especially the ones without distinct foregrounds, like grids. As we expected, our proposed TransMatting model could predict more precise alpha values of highly transparent objects compared with other SOTA methods.

\begin{table}[t]
\centering
\caption{The quantitative results on Distinctions-646 test set.
% $^{\ast}$ denotes methods with tri-token introduced in the convolutional networks. 
% 
}
\label{tab:sota:Distinctions}
\begin{tabular}{@{}c|cccc@{}}
\toprule
Methods       & SAD $\downarrow$   & MSE $\downarrow$  & Grad. $\downarrow$  & Conn. $\downarrow$  \\ \midrule 
KNNMatting \cite{chen2013knn}    & 116.68 & 25   &103.15 & 121.45 \\
DIM \cite{xu2017deep}           & 47.56 & 9    & 43.29 & 55.90 \\
HAttMatting \cite{qiao2020attention}   & 48.98 & 9    & 41.57  & 49.93 \\
% Context-Aware Matting\cite{hou2019context}$^*$ & 36.32 & 7.1  & 29.49 & 35.43 \\
GCAMatting \cite{li2020natural}    & 27.43 & 4.8  & 18.7 & 21.86 \\ 
MGMatting \cite{yu2021mask} & 33.24 & 4.51 & 20.31 & 25.49 \\ \midrule 
TransMattingV1 & 25.65 & 3.4     & 16.08      & 21.45       \\ 
TransMattingV2 & \textbf{24.19} & \textbf{3.0}     & \textbf{14.42}      & \textbf{19.14}       \\ \bottomrule 
\end{tabular}
\end{table}

\begin{figure*}[t]
\centering
\includegraphics[width=0.9\textwidth]{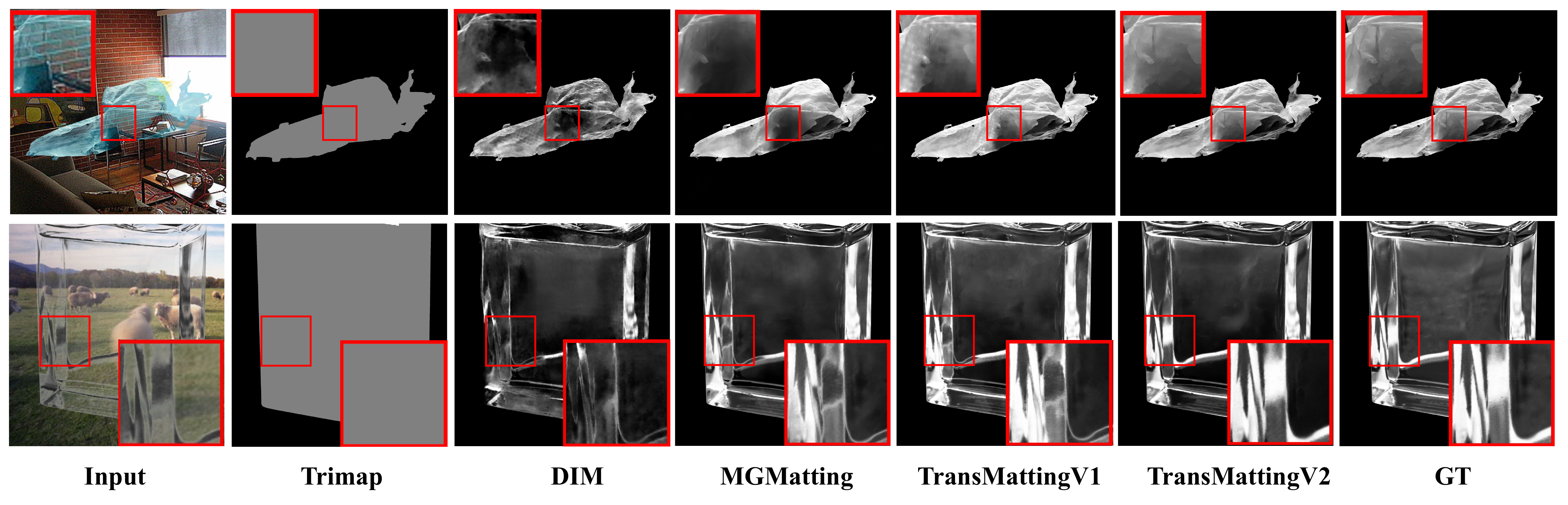}
\caption{Visual comparison of our TransMatting against SOTA methods on our \Ourds \ test set.}
\label{fig:Transpaernt-460-test-set}
\end{figure*}
% \vspace{-16pt}

\begin{table}[t]
\centering
\caption{The quantitative results on our proposed \Ourds{} test set. 
% $^{\ast}$ denotes methods with tri-token introduced in the convolutional networks.
}
\label{tab:sota:ours}
\begin{tabular}{c|cccc}
\toprule
Methods       & SAD$\downarrow$   & MSE$\downarrow$ & Grad.$\downarrow$  & Conn.$\downarrow$  \\ \midrule
IndexNet \cite{lu2019indices}      & 573.09 & 112.53  & 140.76 & 327.97 \\
MGMatting \cite{yu2021mask}    & 111.92 & 6.33  & 25.67 & 103.81 \\ \midrule 
TransMattingV1 & 88.34      & \textbf{4.02}    & 20.99      & 82.56     \\
TransMattingV2 & \textbf{87.26}      & 4.06    & \textbf{20.86}      & \textbf{80.6}     \\
\bottomrule
\end{tabular}
\end{table}

\textbf{Testing on our \Ourds.}
Based on their release codes, we train IndexNet and MGMatting methods on our dataset and compare them with ours in Tab.~\ref{tab:sota:ours}. Our \Ourds\ dataset mainly focuses on transparent and non-salient foregrounds, which is very difficult for existing image matting methods. Surprisingly, as illustrated in Tab.~\ref{tab:sota:ours}, our TransMattingV1 achieves a promising result with only 4.02 MSE and surpasses other methods by a large margin on SAD, Grad., and Conn. In addition, our TransMattingV2 has further improved compared to TransMattingV1, especially on Conn.
Furthermore, to evaluate the generalization performance of our model, we train our TransMatting on the Composition-1k training set and directly test it on the \Ourds\ test set. The results are shown in Tab.~\ref{tab:generalize-our}. Thanks to the large receptive field and well-designed multi-scale fusion module, our model reduces more than half of the SAD, MSE, and Conn. As shown in Fig.~\ref{fig:Transpaernt-460-test-set}, we visualize the comparison results of two representative TT-type objects, a plastic bag and a glass bottle. We can see that our TransMatting shows more sophisticated and accurate alpha mattes than other methods. In short, both quantitative and generalization comparisons prove the effectiveness and superiority of our proposed method.

\begin{table}
\centering
\caption{Generalization results on our proposed \Ourds{}   test set.
% $^{\ast}$ denotes methods with tri-token introduced in the convolutional networks.
}
\label{tab:generalize-our}
\begin{tabular}{c|cccc}
\toprule
Methods       & SAD$\downarrow$   & MSE$\downarrow$ & Grad.$\downarrow$  & Conn.$\downarrow$  \\ \midrule
DIM \cite{xu2017deep}       & 351.98   & 53   & 151.37   & 292.04  \\
IndexNet \cite{lu2019indices}      & 434.14  & 74.73   & 124.98  & 368.48\\
MGMatting \cite{yu2021mask}    & 344.65  & 57.25   & 74.54    & 282.79\\
TIMI-Net \cite{liu2021tripartite}  & 328.08      &  44.2    & 142.11     & 289.79      \\ \midrule 
TransMattingV1 & 192.36     & 20.96    &  41.8     & \textbf{158.37}      \\ 
TransMattingV2 & \textbf{181.94}      & \textbf{18.37}    &  \textbf{40.99}     & 159.53     \\
\bottomrule 
\end{tabular}
\end{table}

\begin{table}[t]
\centering
\caption{Generalization results on AIM-500. All the models are
% the officially provided model, which are merely 
trained with the Composition-1k training set.  }
\label{tab:generalize-AIM500}
\begin{tabular}{c|cccc}
\toprule 
Methods       & SAD $\downarrow$   & MSE $\downarrow$ & Grad. $\downarrow$  & Conn. $\downarrow$  \\ \midrule 
IndexNet \cite{lu2019indices}      & 28.49 & 28.8 &18.15  &27.95 \\
CA \cite{hou2019context}                                & 26.33 & 26.6 &18.89  &25.05 \\
CA+DA  \cite{hou2019context}                            & 32.15 & 38.8 &30.25  &31.00 \\
GCAMatting \cite{li2020natural}                        & 35.10 & 38.9 &25.67  &35.48 \\
$A^2$U \cite{dai2021learning}                             & 30.38 & 30.7 &22.60  &30.69 \\
SIM \cite{sun2021semantic}                            & 27.05 & 31.1 &23.68  &27.08 \\ \midrule 
%FBA                             & 19.05 & 16.2 &11.42  &18.30 \\ 
TransMattingV1 & 16.55      & 14.06    &  12.35 & 15.61 \\ 
TransMattingV2 & \textbf{15.70}      & \textbf{12.20}    &  \textbf{11.36} & \textbf{15.05} \\

\bottomrule
\end{tabular}
\end{table}

\subsubsection{Experiments on Real-World Dataset}
\textbf{Generalization on AIM-500.} 
Besides synthetic benchmarks, we also validate our method on real-world images, which is more challenging than synthetic ones. Specifically, we train the network on the synthetic Composition-1k training set and directly evaluate on AIM-500.
The results are shown in Tab.~\ref{tab:generalize-AIM500}. 
As we can see, without any extra data augmentation or fine-tuning, our model surpasses other SOTA methods with a big margin.
The performance of TransMattingV2 is better than TransMattingV1, thanks to the introduced tri-token in the convolutional network. Furthermore, AIM-500 also has some real-world transparent objects, like glass, fire, plastic bags, etc. 
Although other methods fail to process such difficult cases, our model still has superior performance compared with them on the generalization results, which further proves the effectiveness of our model on transparent objects.

\begin{figure}
\centering
\includegraphics[width=0.9\linewidth]{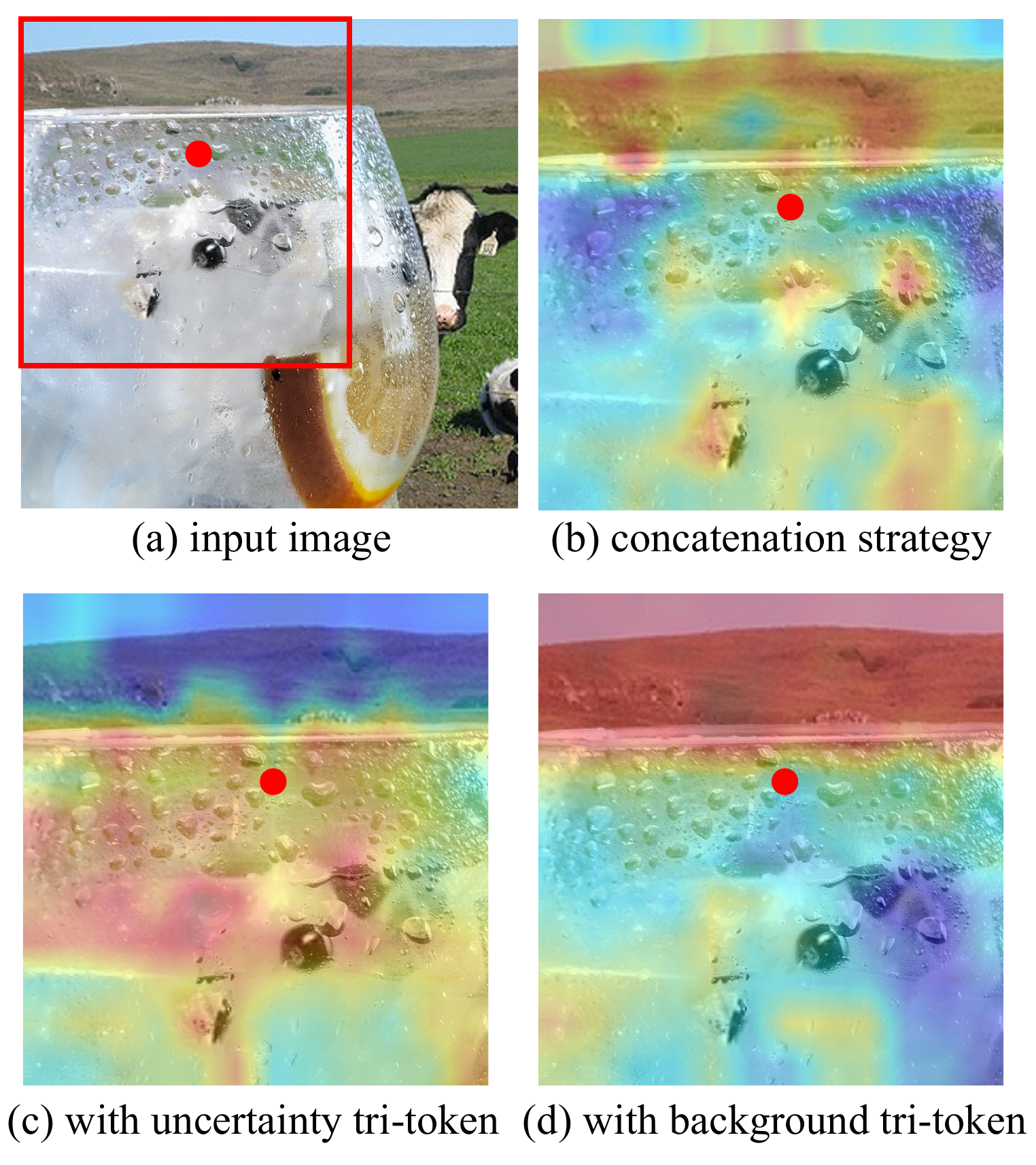}
\caption{Visualization of attention maps ($q @ k$) of the self-attention mechanism. The color indicates the attention weight between the red point and other image pixels. (a) is the input image, (b) and (c) represent the attention maps of the traditional concatenation strategy and our tri-token manner. In (d), the tri-token values are replaced from uncertainty to background values, and the model is deceived into paying attention to the background, indicating the guiding efforts of our tri-token. }
\label{fig:attention-map}
\end{figure}

\section{Analysis of the Tri-token}
The tri-token is proposed to effectively introduce semantic and position information in trimap to deep features. We hypothesize that this could help the model find where it should pay more attention. In Fig.~\ref{fig:attention-map}, we visualize attention maps of the self-attention mechanism according to the red-marked point. As the red point belongs to the transparent glass cup, the pixels belonging to the cup should have a high weight. However, as illustrated in Fig.~\ref{fig:attention-map} (b), many background pixels are highly activated with the concatenation strategy. While with our tri-token (shown in Fig.~\ref{fig:attention-map} (c)), the background is suppressive, and the model mainly concentrates on the foreground, indicating that our tri-token can efficiently guide the model to distinguish different areas. To further demonstrate the effectiveness of the tri-token, we replace the token in the red point with background values. As we can see from Fig.~\ref{fig:attention-map} (d), the replaced tri-token misleads the model to mistakenly focus on the background pixels with very high response, meaning that the model relies on the guidance from our tri-token to distinguish foreground and background while traditional concatenation loses much information in deep layers.

\section{Conclusion}
In this paper, we present a Transformer-based network for image matting (TransMatting) to mine long-range features of foreground objects, especially for transparent and non-salient ones. A multi-scale global-guided fusion module is proposed to take the global information as a guide to fuse multi-scale features for better modelling the uncertainty regions in transparent objects. We also collect a high-resolution matting dataset of transparent objects to serve as a test bed for future research. Furthermore, we redesign the trimap as a unified representation to directly introduce semantic information to deep features instead of traditional concatenation. Various experiments demonstrate that it could boost both Transformer and convolutional networks, indicating that the proposed tri-token has the potential to be a new paradigm for image matting.

% or


% \appendices

% % you can choose not to have a title for an appendix
% % if you want by leaving the argument blank
% \section{}

% % use section* for acknowledgment
% \ifCLASSOPTIONcompsoc
%   % The Computer Society usually uses the plural form
%   \section*{Acknowledgments}
% \else
%   % regular IEEE prefers the singular form
% \fi


\ifCLASSOPTIONcaptionsoff
  \newpage
\fi



\bibliographystyle{IEEEtran}

%\begin{thebibliography}{1}
%H.~Kopka and P.~W. Daly, \emph{A Guide to {\LaTeX}}, 3rd~ed.\hskip 1em plus
%\end{thebibliography}

% \bibliographystyle{ECCV}
\bibliography{egbib}



%\newpage



\vfill


\end{document}